# A Software-Only Post-Processor for Indexed Rotary Machining on GRBL-Based CNCs

**Pedro Portugal [1], Damian Venghaus [2] and Diego Lopez [3,*]**

[1] Tecnologico de Monterrey, Queretaro, 76130, Mx. Independent Researcher.
[2] Grafisch Lyceum Rotterdam, Rotterdam, 3013 AK, Nl. Independent Researcher.
[3] Tecnologico de Monterrey, Queretaro, 76130, Mx. Independent Researcher.
* Correspondence: pedroportugalsil@gmail.com

**Abstract**

Affordable desktop CNC routers are widely used in education, prototyping, makerspaces, and even small-scale manufacturing but most lack a rotary axis, limiting their ability to fabricate rotationally symmetric or multi-sided components. Existing solutions often require hardware retrofits, alternative controllers, or commercial CAM software, increasing cost and complexity. This work introduces a software-only framework for indexed rotary machining on GRBL-based CNC machines. The framework combines a custom post-processor, which converts planar toolpaths into discrete rotary steps, with a browser-based interface for execution. While not equivalent to continuous 4-axis machining, the method enables reliable rotary-axis fabrication using only common, off-the-shelf mechanics and without firmware modification. Experiments demonstrate the framework's ability to accurately produce rotationally symmetric components in a variety of materials. By lowering technical and financial barriers, this approach broadens access to multi-axis machining in classrooms, makerspaces, and small workshops, supporting hands-on learning and rapid prototyping. The results indicate that software-only solutions can provide a practical and cost-effective alternative for small-scale rotary machining applications.

**Keywords:** CNC; GRBL; indexed rotary machining; post-processor; multi-axis fabrication; rapid prototyping; makerspaces; decentralized manufacturing.

## 1. Introduction

Advances in manufacturing technologies—from the mechanization of the Industrial Revolution to contemporary FabLabs and desktop CNC machines—have progressively lowered barriers to designing and producing complex artifacts [1]. Affordable desktop CNC routers are widely used in education, prototyping, makerspaces and even small-scale manufacturing, yet most lack a rotary axis, limiting fabrication of rotationally symmetric or multi-sided components [2]. Commercial 4-axis desktop machines typically cost between USD $5,000–$10,000, compared to less than $1,000 for GRBL-based systems, making multi-axis machining financially inaccessible for most educational and small-scale contexts [3-5].

GRBL, an open-source firmware for Arduino-based CNC platforms, provides reliable planar motion but lacks native rotary-axis support (at least on the Vanilla version), restricting production of shafts, gears, and cylindrical models. Existing solutions either rely on hardware retrofits, alternative controllers, or commercial CAM software, increasing cost and complexity [6–9]. Software-driven methods can emulate rotary motion at the G-code level using inexpensive, off-the-shelf components, but no lightweight, software-only approach preserves the low-cost, low-complexity advantages of GRBL [10,11].

This work introduces a **post-processing framework** that injects indexed rotary commands into conventional G-code, enabling low-cost GRBL-based CNC machines to approximate 4-axis functionality without modifying firmware. The method converts standard planar toolpaths into segmented toolpaths interleaved with rotary steps, using only commodity mechanics. By lowering technical and financial barriers, it broadens access to multi-axis machining for education, makerspaces, and small-scale manufacturing. The framework supports hands-on learning, rapid prototyping, and community-based fabrication, demonstrating that software-only solutions can extend the capabilities of low-cost CNC systems.

## 2. Materials and Methods

### *2.1 Toolpath Design*

Standard computer-aided manufacturing (CAM) tools were used to generate initial toolpaths for the proposed method. Commercial software packages such as CAMWorks, Mastercam, or Fusion 360 are fully compatible with the developed post-processor. In this work, Fusion 360 was selected to illustrate toolpath generation and structure. Solids of revolution were chosen as test parts due to their prevalence in mechanical components and suitability for demonstrating discrete rotary indexing.

The workflow begins with importing a CAD model into the CAM environment, followed by preprocessing to ensure compatibility with the post-processor. The part is oriented such that the longitudinal axis aligns with X and the radial axis with Z, then reduced to a 2D XZ cross-sectional profile (Figure 1). Profile thickness must remain below the cutting tool diameter to keep all tool motion within the XZ plane and ensure the toolpath contains only one pass.

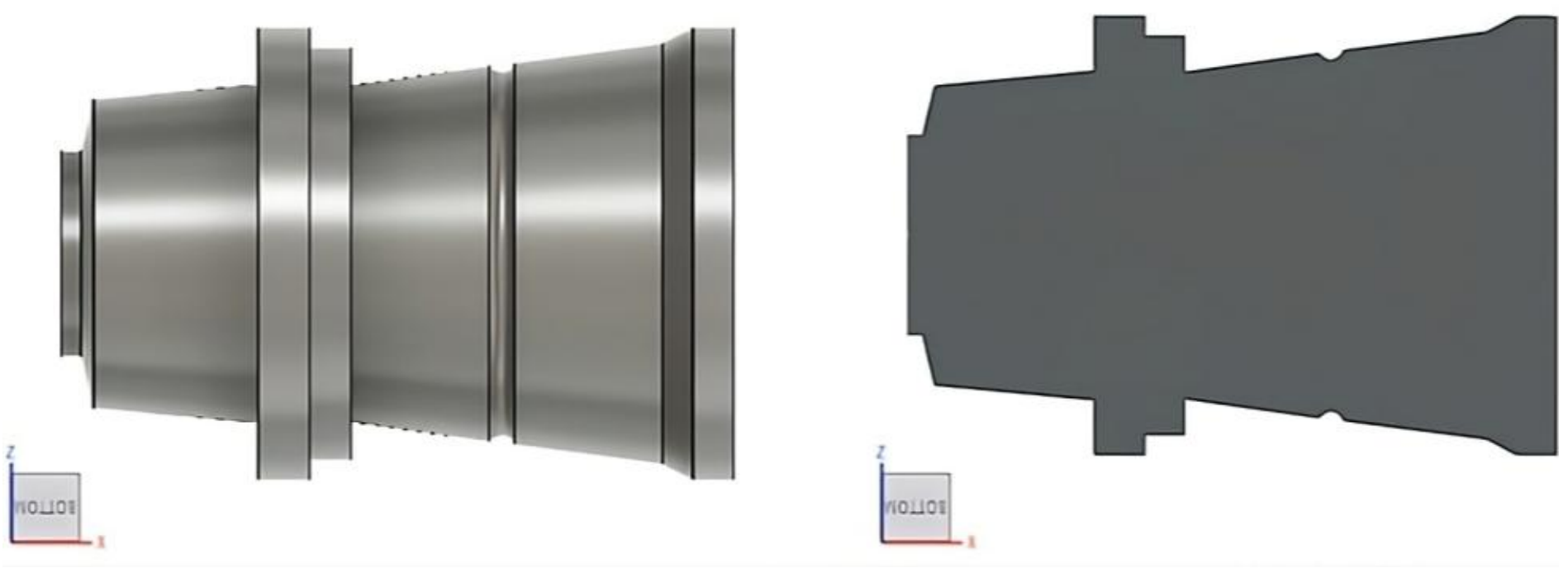

***Figure 1.*** *CAD model of the part to be manufactured. The original revolved geometry (left) and its cross-sectional representation (right), both oriented along the XZ plane.*

A 3D parallel machining strategy was employed to generate toolpaths due to its consistent engagement and predictable cutting loads [12]. Feed rate, spindle speed, and step-over were reduced to approximately 75% of conventional milling values to compensate for discrete indexing. Once generated, the toolpath conforms to the outline of the XZ cross-section (see Figure 2). The post-processor then interprets this planar motion and injects synchronized Y-axis commands to emulate rotary motion.

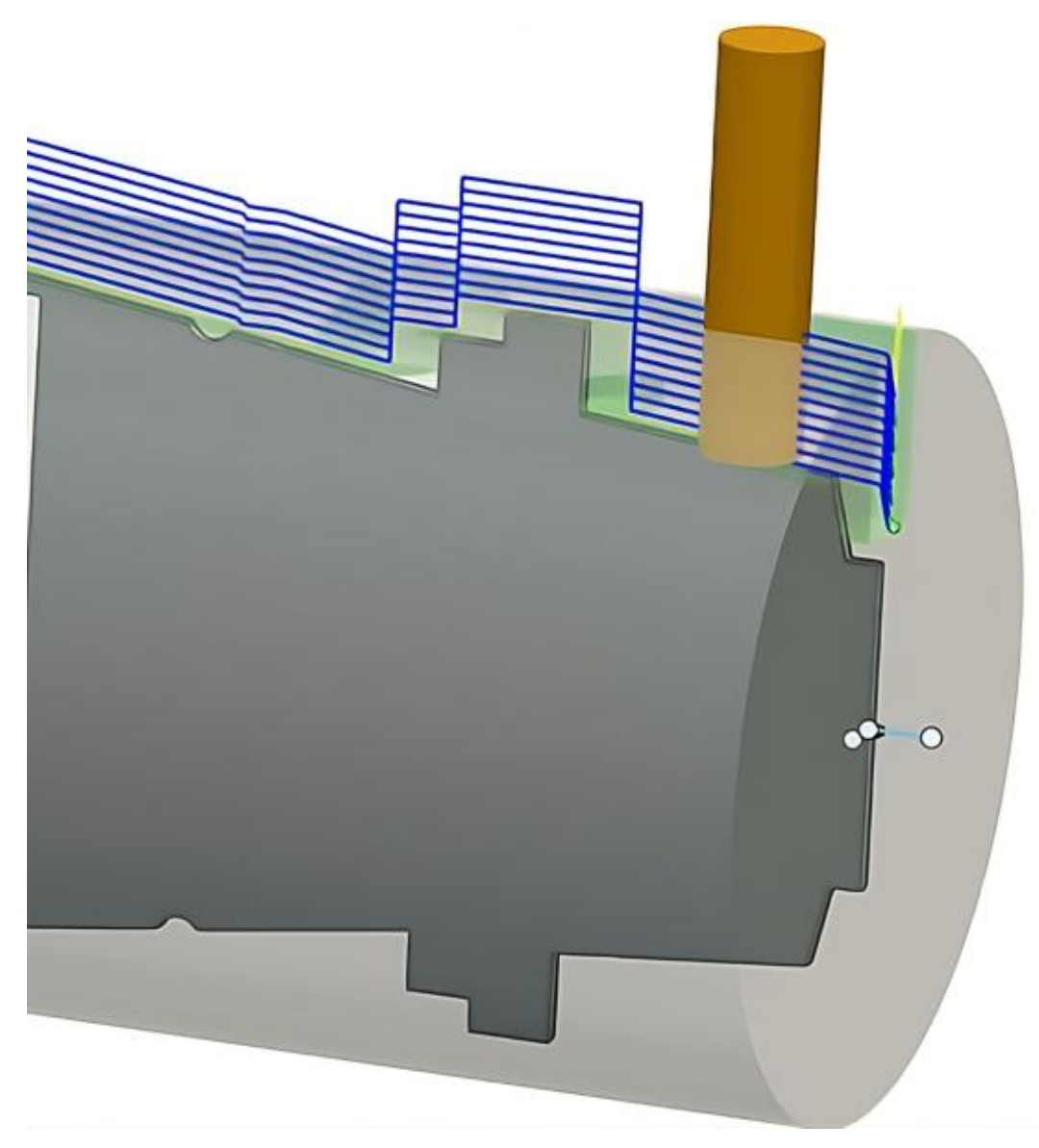

***Figure 2.*** *CAM simulation showing a 3D parallel toolpath along the XZ cross-section contour.*

## 2.2 Post-Processor Design

The post-processor was developed in Python to convert planar XZ toolpaths into discrete rotary motion. While prior work has explored post-processing for multi-axis CNCs [13,14], this tool uniquely applies discrete indexing logic to emulate a fourth axis on standard three-axis hardware. The tool operates as both a command-line utility and a GUI, supporting batch and interactive workflows.

This approach is analogous to intermittent indexing methods used in high-end rotary systems where rotational motion is divided into discrete steps, such as those described in recent studies on bi-rotary milling heads [15]

$$w = \alpha D_{tool} \tag{1}$$

$$N = \left[\frac{\pi D_{Stock}}{w}\right] \tag{2}$$

$$\theta = \frac{360^\circ}{N} \tag{3}$$

Where w is the effective cut width, $\alpha$ the overlap factor, $D_{tool}$ the tool diameter, $D_{Stock}$ the stock diameter, and θ the angular step per pass. Optional calculation of maximum radial deviation $e$ allows control of surface finish:

$$e = R_{Stock}\left(1 - cos\left(\frac{\theta\pi}{180^\circ}\right)\right) \tag{4}$$

$$e \approx \frac{R_{Stock}\,\pi^2}{2N^2}, \quad N \approx \pi\sqrt{\frac{R_{Stock}}{2e}} \tag{5,6}$$

Each pass begins with a G0 Yθ command, followed by the original XZ toolpath, with spindle stopped and Z-axis retracted during indexing. The post-processor includes automatic parameter extraction from G-code comments and a visualization module for validating toolpaths prior to execution (Figure 3).

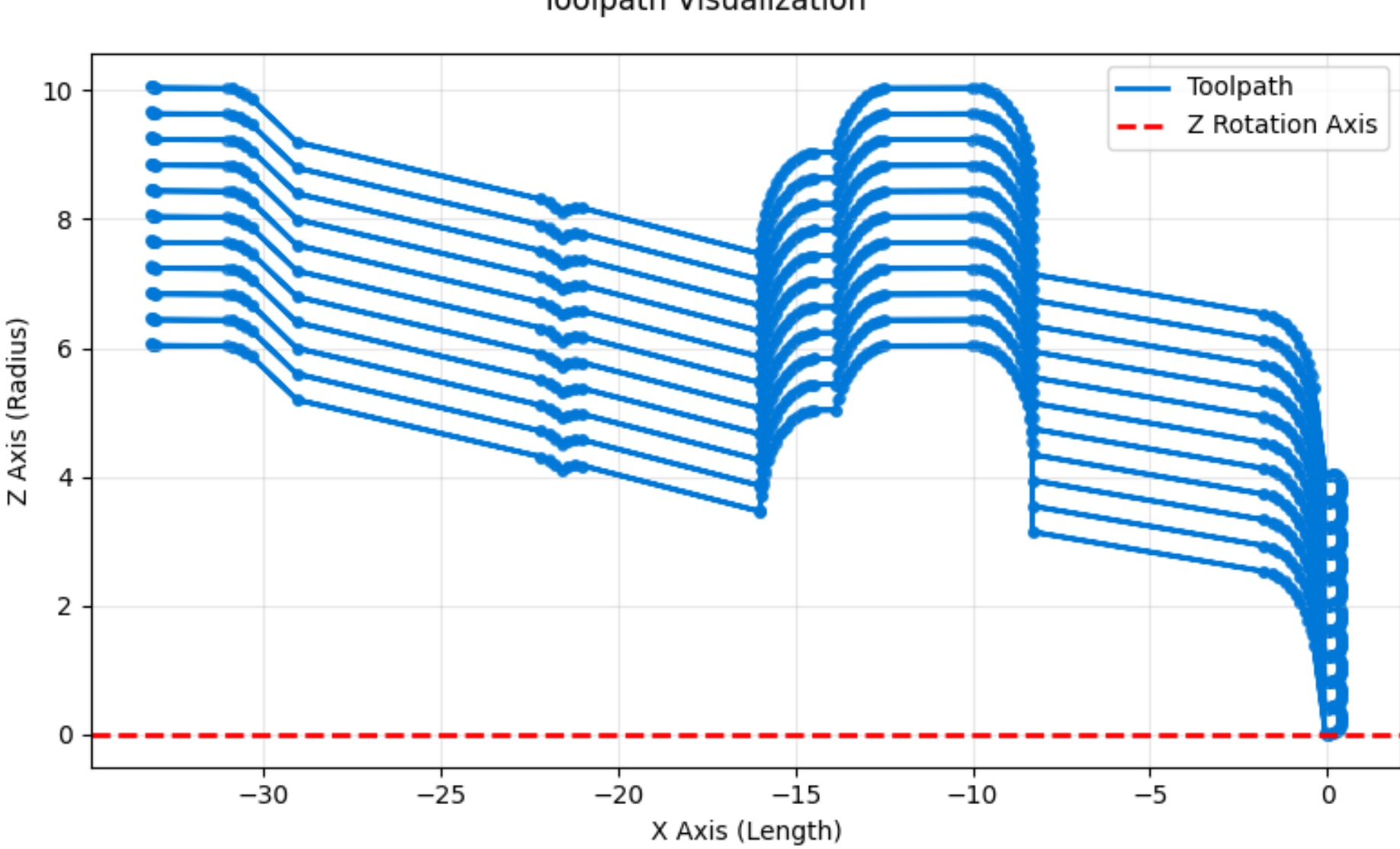

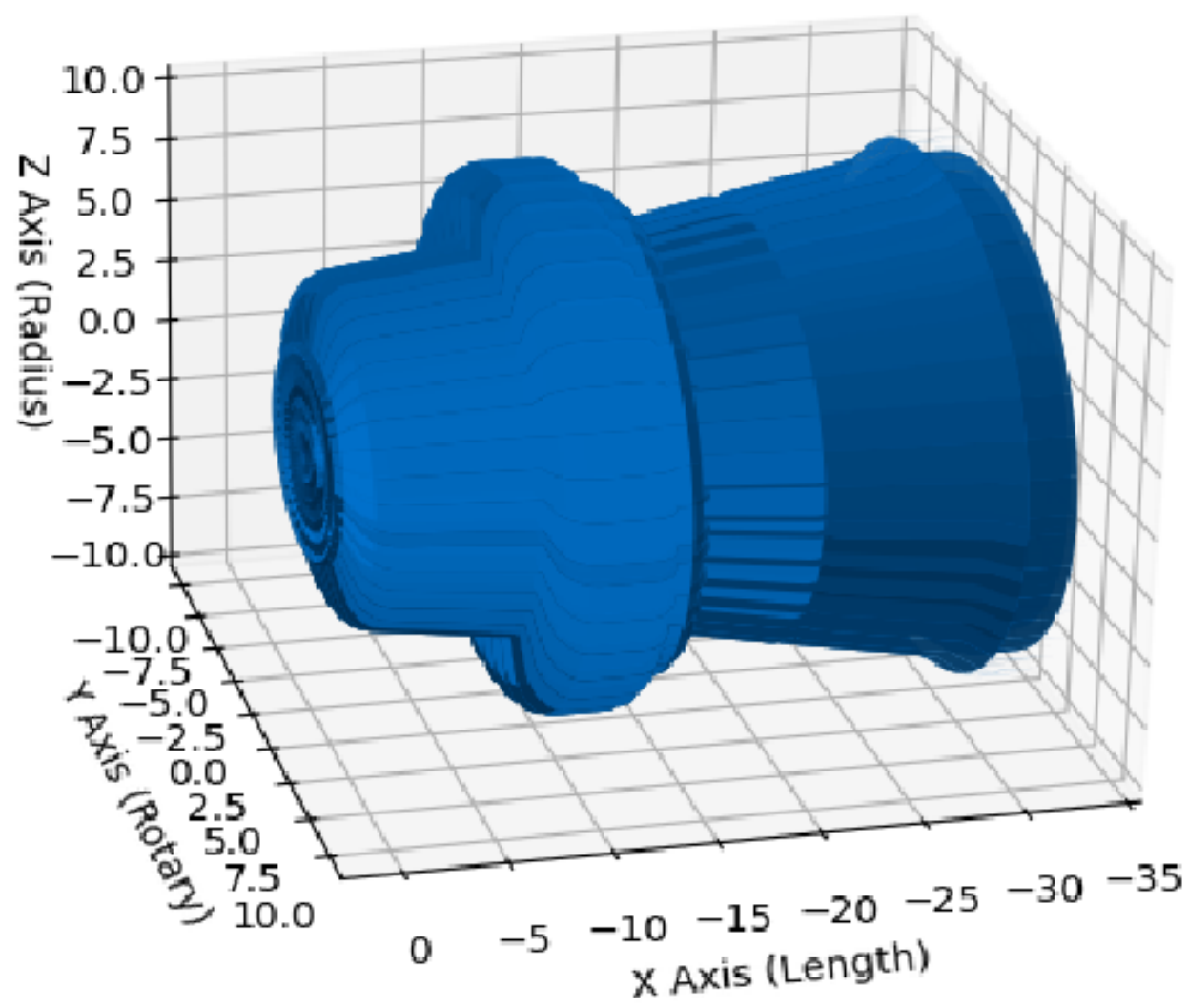

***Figure 3.*** *Visualization of the geometry to machine extracted from the post-processed G-code.*

## *2.3 Hardware Configuration*

Rotary motion was implemented on a GRBL-based CNC via a mechanical assembly consisting of a 2-phase NEMA 23 stepper motor, timing belt and pulley system, and 63 mm 3-jaw chuck mounted on a rigid aluminum fixture (Figure 4). The pulley transmission provides mechanical decoupling between motor and machining forces, allowing easy adjustment of gear ratios. Bearings support smooth rotation of the stock.

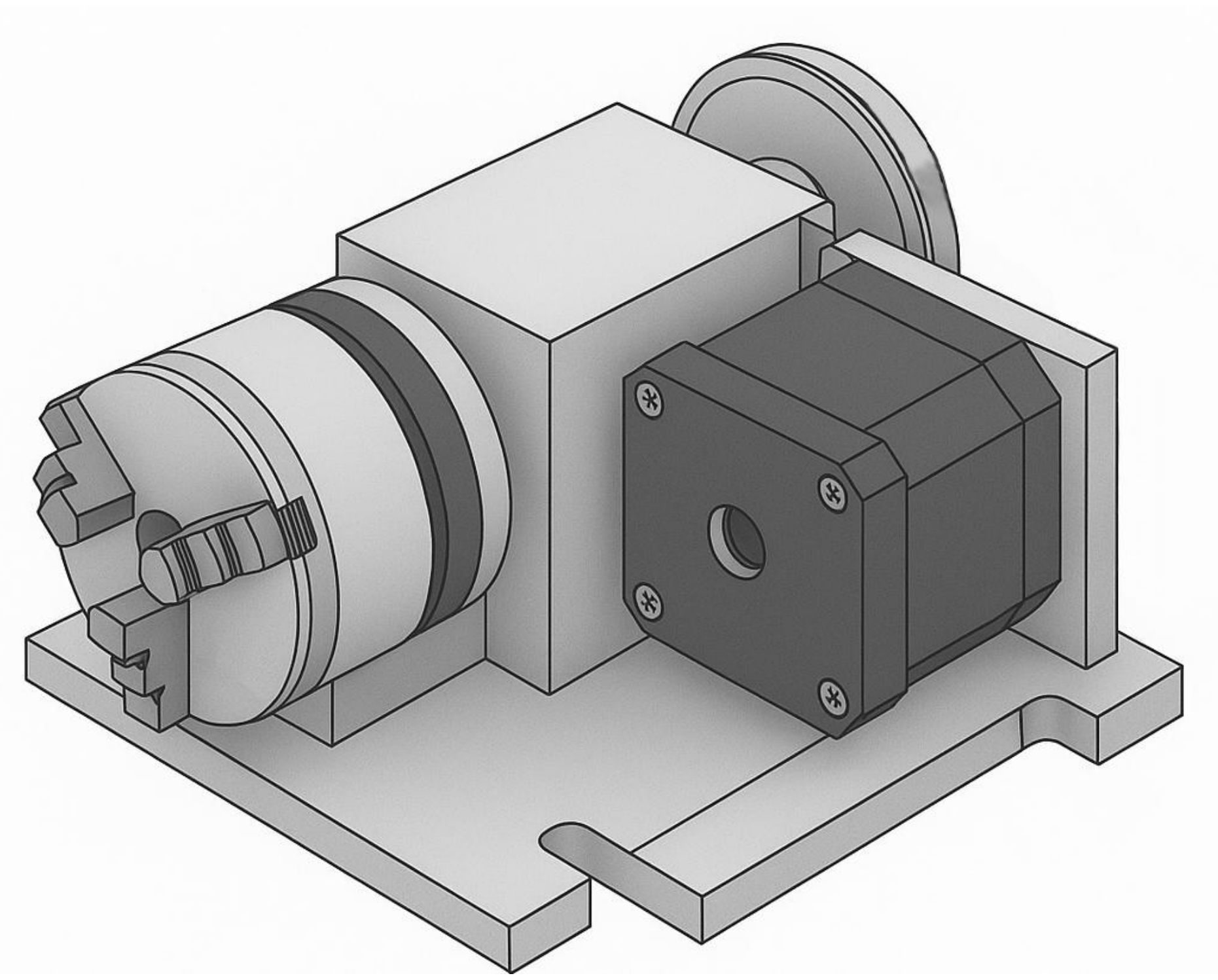

***Figure 4.*** *Isometric view of the rotary axis assembly used on the CNC machine, showing the 2-phase NEMA*

The assembly is modular, enabling detachment for standard 3-axis operation. A 1:1 pulley ratio was used to maintain direct correspondence between motor steps and rotary increments calculated by the post-processor. Non-1:1 ratios require modification of post-processor scaling. Users can adapt the assembly to different machines or constraints, as long as stepper motion can be interfaced with the control board, where the rotary assembly is to be connected to the Y-axis output.

Safety protocols—including spindle stop, Z-axis retraction, and controlled restarts—are automatically applied during indexing to prevent collisions and ensure reliable operation.

## 3. Results

### *3.1 Post-Processor Implementation*

The post-processing tool was developed in Python and follows a modular architecture organized into three subsystems: (a) G-code parsing and modification, (b) graphical user interface (GUI), and (c) visualization and simulation module. Python's text-processing capabilities and scientific libraries make it well suited for CNC post-processing applications [16,17].

The system is compatible with standard GRBL-style G-code and requires no firmware modifications, assuming only the presence of a rotary chuck and stepper-driven axis. The core logic parses the input G-code line by line, extracts X and Z coordinates, removes pre-existing Y commands, and prepares repeated execution at discrete angular intervals. Each pass begins with a G0 Yθ command for angular indexing, followed by the original XZ toolpath. This ensures planar motion during each pass while emulating rotary motion across the part.

The built in GUI allows users to specify parameters, preview 2D toolpaths, and visualize a 3D reconstruction of the revolved geometry prior to execution. Metadata such as tool diameter and feedrate are automatically extracted from structured G-code comments, reducing configuration errors. This interactive approach aligns with modern CNC interfaces designed for rapid parameter tuning and on-screen feedback, such as those supported in STEP-NC simulations [18].

The final output is fully GRBL-compatible G-code with indexed rotary commands. A selected excerpt of the code is provided in Appendix A, and the full repository is publicly available for reproducibility.

### 3.2 Web Interface

A browser-based implementation provides a platform-independent interface for planar-to-rotary G-code conversion. Users upload XZ-plane G-code files and configure rotary parameters, including stock radius, rotary resolution, and optional feedrate overrides. Real-time validation ensures numerical integrity and prevents configuration errors.

After processing, the interface generates a modified G-code file with indexed Y-axis motion and displays a 3D preview of the revolved geometry alongside the original 2D toolpath (Figure 5). This enables verification prior to execution. The web interface executes file parsing server-side in ephemeral containers, discards uploaded files after conversion, and enforces a default 5 MB file-size limit to ensure security and portability (see Appendix B).

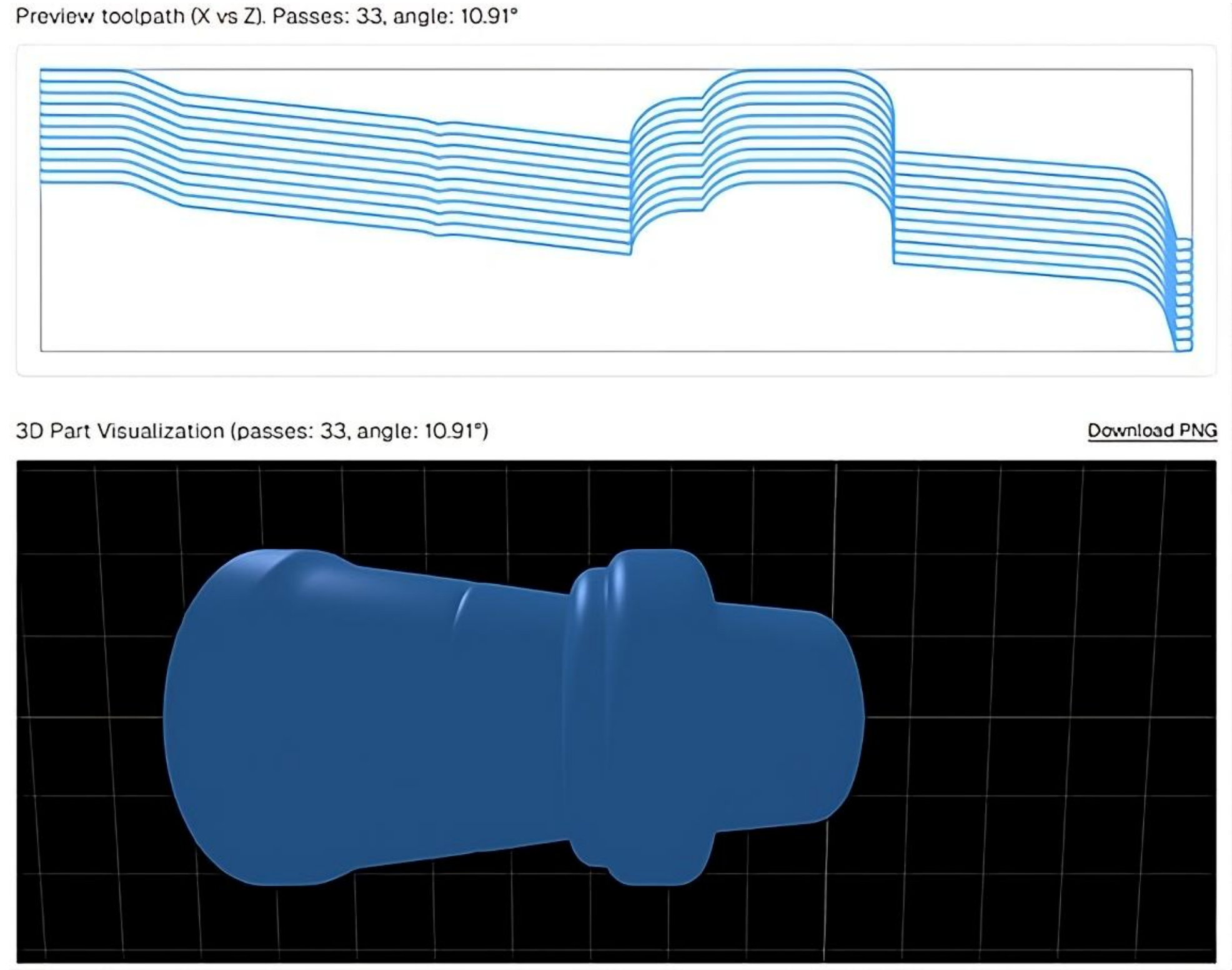

***Figure 5.*** *Web interface of the rotary G-code converter showing the original 2D XZ toolpath (top) and the corresponding 3D reconstruction of the revolved solid (bottom), enabling users to verify toolpaths prior to G-code export.*

### 3.3 Validation and Test Results

The method was validated using a series of test parts machined from hardwood and copper. All parts used the same 2D toolpath generated in Fusion 360 and post-processed through the GUI interface. Machining was performed on a generic 3020 CNC router equipped with a 500W air-cooled spindle and a 3.175 mm HSS two-flute flat end mill. Feedrates and spindle speeds were adjusted per material (wood: 1,300 mm/min at 8,000 RPM; copper: 500 mm/min at 12,000 RPM).

For a stock diameter of 22 mm, the system computed N = 80 indexed passes with an angular step of approximately 4.5°. Dimensional validation with digital calipers showed the wooden part measured 20.30 mm (−0.20 mm error) and copper 20.75 mm (+0.25 mm error), consistent with expected tolerances from tool deflection and material properties.

Figures 6–9 illustrate the resulting parts: wooden output (Fig. 6), copper output (Fig. 7), side-by-side comparison (Fig. 8), and copper toolmark close-up (Fig. 9) demonstrating discrete angular indexing behavior.

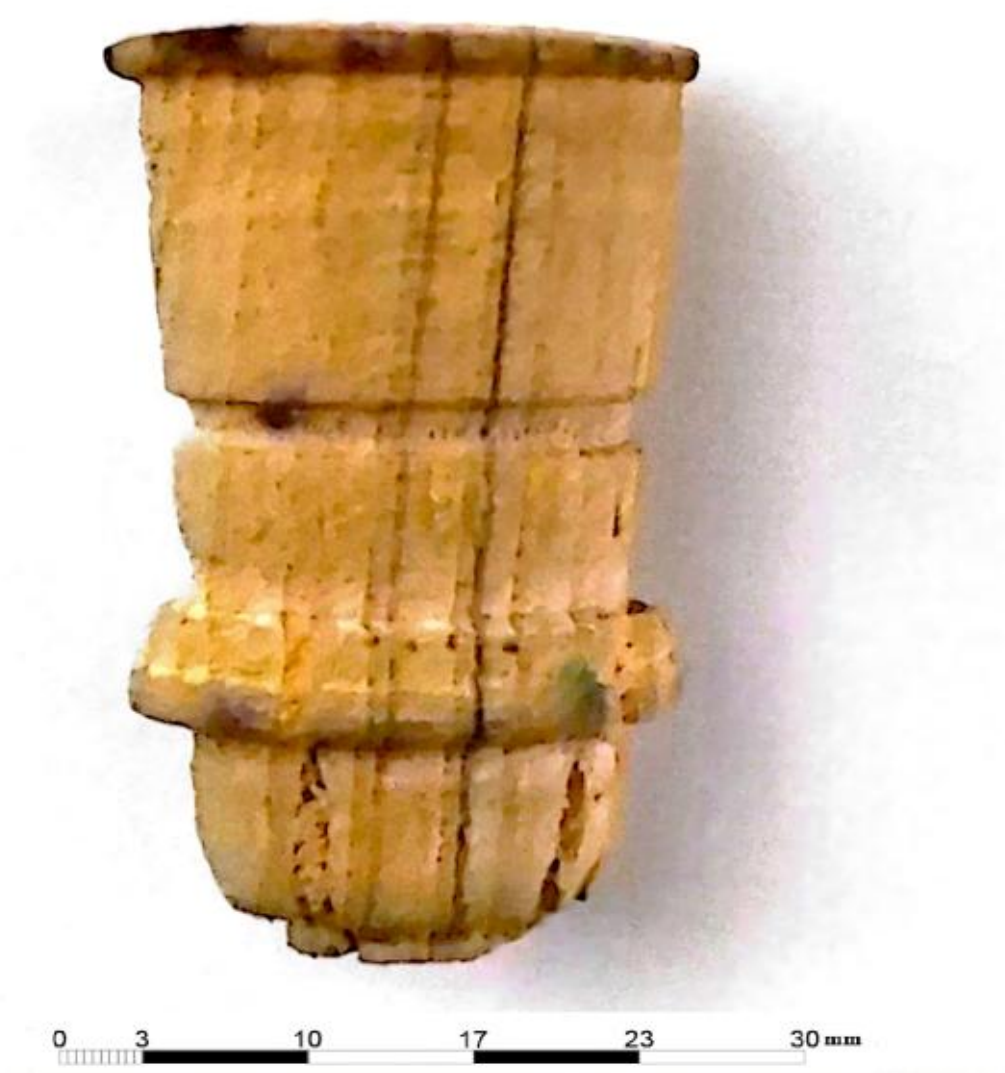

***Figure 6.*** *Machined test part in hardwood using the proposed indexing method.*

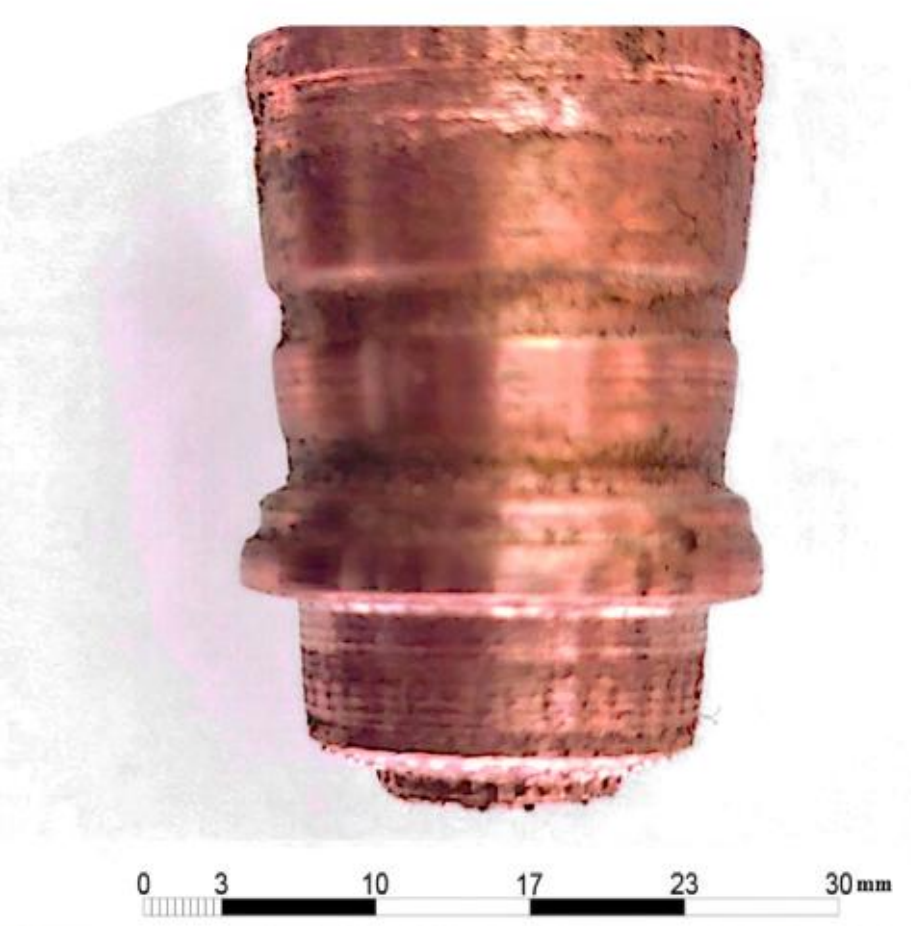

***Figure 7.*** *Copper version of the sample geometry, machined with coolant and adjusted parameters.*

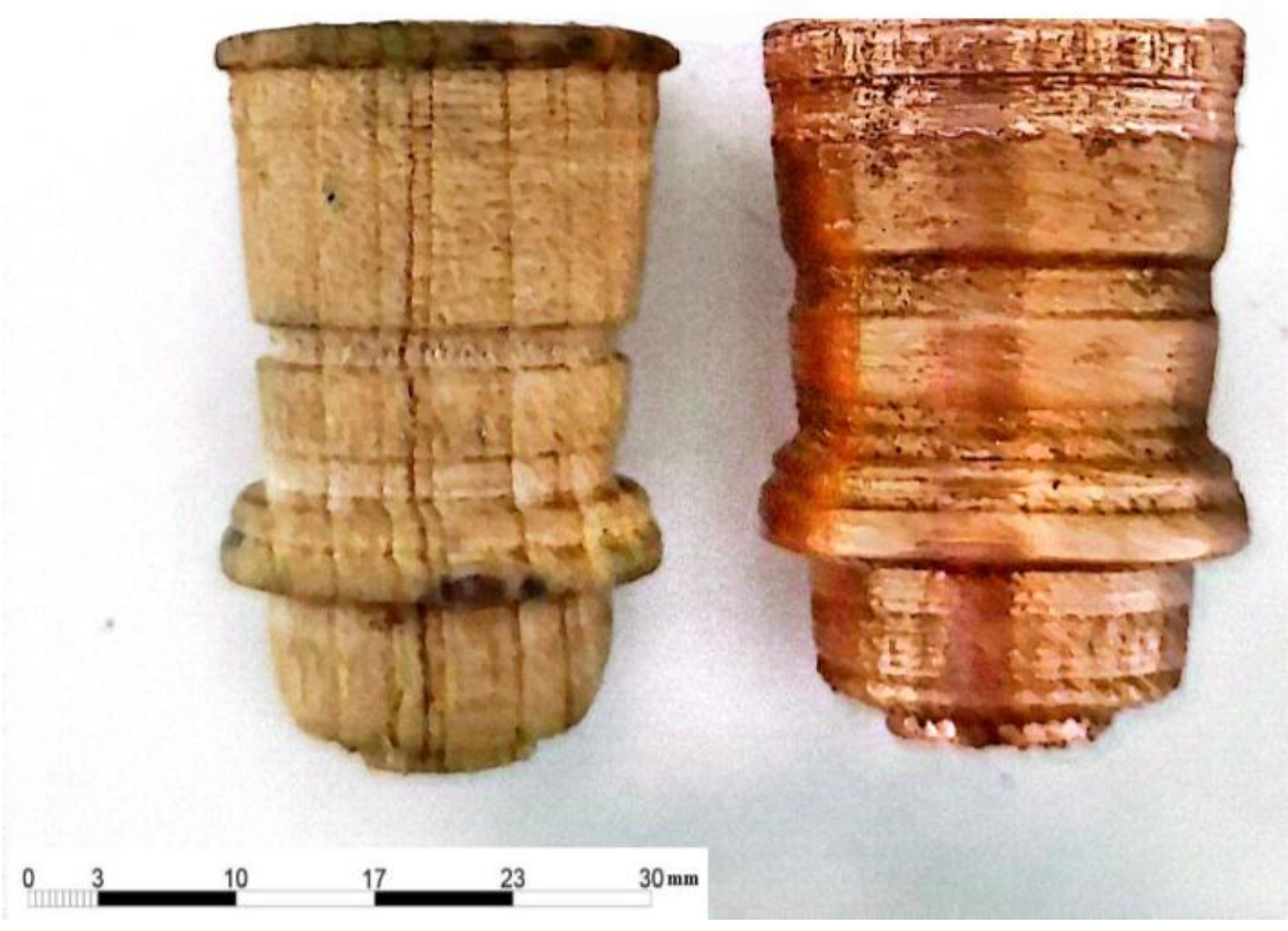

***Figure 8.*** *Side by side comparison of wood and copper parts, showing material related differences.*

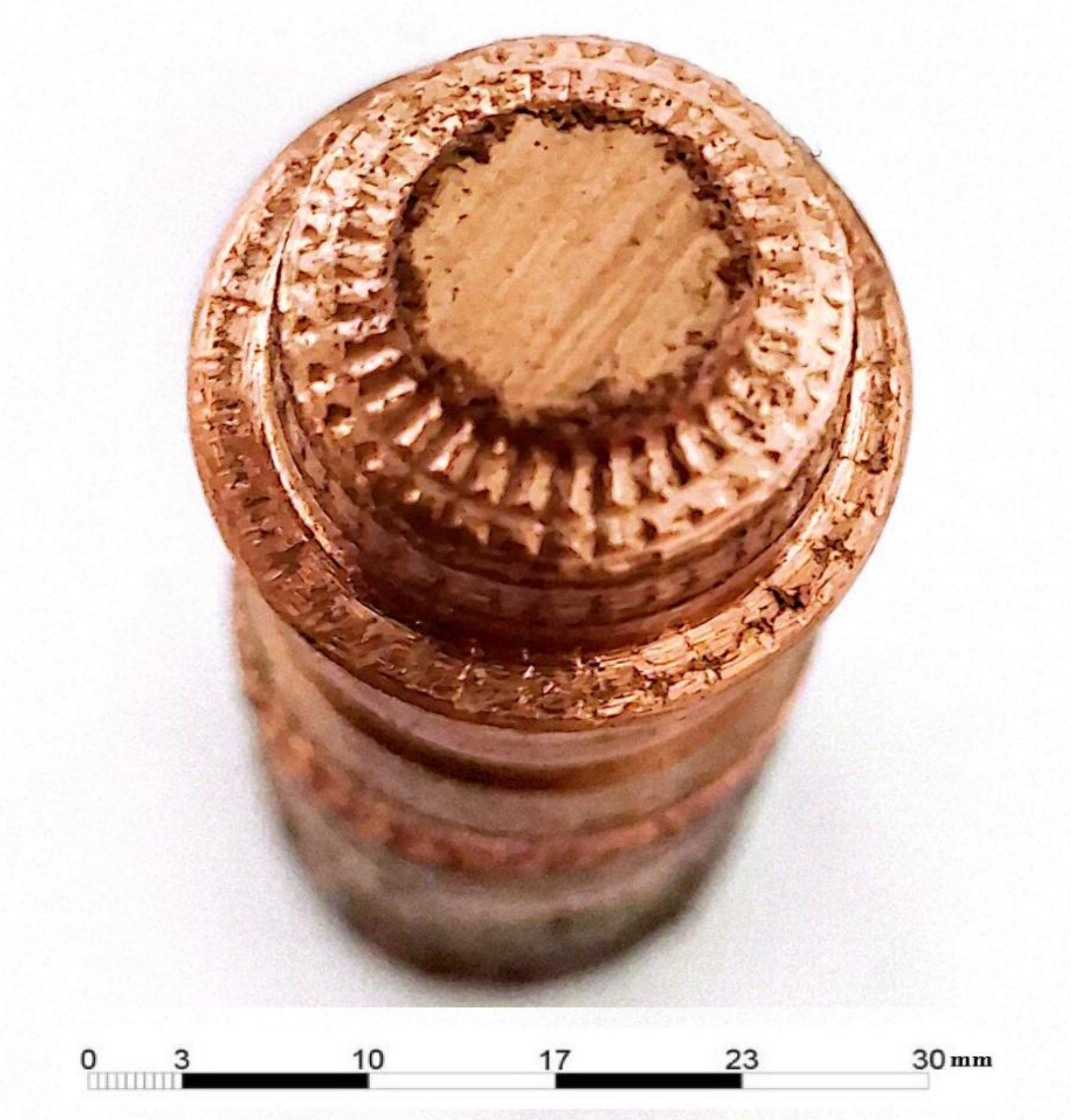

***Figure 9.*** *Close-up of copper base showing concentric toolmarks from discrete angular indexing.*

Figure 10 contrasts parts produced with coarse versus fine angular steps, showing the trade-off between surface fidelity and machining time.

These results confirm the post-processor reliably generates indexed rotary commands, produces accurate parts, and is compatible with low-cost GRBL-based CNC systems. Users reported that the 3D preview facilitated verification and minimized errors, demonstrating the usability and practical applicability of both the desktop GUI and web interface.

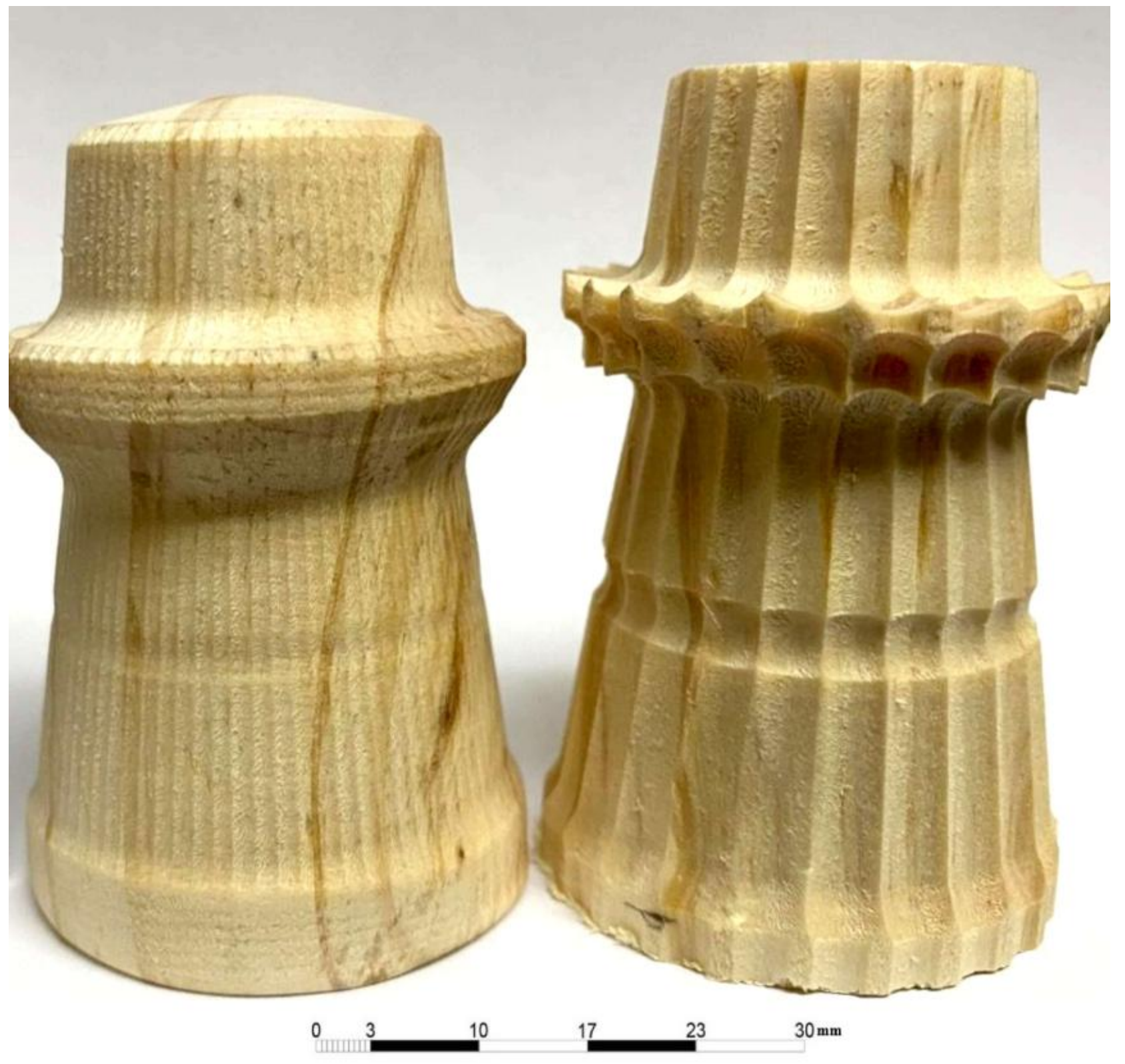

***Figure 10.*** *Side-by-side comparison of coarse $N=22$ (right) and fine $N=80$ (left) indexing. Coarse indexing exhibits visible indexing artifacts and scalloping; finer indexing improves surface continuity at the cost of longer cycle time and larger programs.*

The machined parts illustrate the practical application of the post-processor; reported deviations (wood −0.20 mm, copper +0.25 mm) are indicative only, as physical results depend on CNC machine, spindle, tooling, and material. The primary validation is the correct injection of indexed rotary commands, accurate angular step calculations, and generation of GRBL-compatible G-code. Hence, a full statistical dataset of variation in the reported results is beyond the scope of this work.

## 4. Discussion

Emulating a fourth axis on low-cost CNC machines broadens access to advanced manufacturing. By translating planar G-code into indexed rotary commands without hardware modifications, this software-driven method circumvents traditional barriers. The results are discussed below in terms of accessibility, trade-offs, comparative performance, and broader implications.

### *4.1 Accessibility and Democratization of Rotary Machining*

Traditional fourth-axis setups require specialized machines or external rotary tables with drivers and CAM support, imposing financial and technical barriers. By operating entirely at the G-code level and re-purposing an unused axis, standard 3-axis machines can simulate 4-axis motion, democratizing multi-axis machining and aligning with open-source fabrication trends.

### 4.2 Technical Trade-Offs and Limitations

Discretized rotation produces visible tool marks and flat facets; smaller angular steps improve surface quality but increase machining time and file size. Torque consistency and dynamic compensation of true rotary axes cannot be replicated. Accuracy depends on machine calibration, and deviations in test parts (wood −0.20 mm; copper +0.25 mm) highlight practical limits.

### 4.3 Comparative Performance with True 4-Axis Systems

This method does not replace industrial fourth axes but offers a favorable cost-performance ratio in educational or low-cost contexts. Discrete indexing simulates quasi-continuous rotation with minimal idle time. Despite minor deviations, parts meet prototyping and decorative requirements without extra drivers or hardware modifications.

### 4.4 Educational and Decentralized Manufacturing Implications

The approach enables safe, low-cost exploration of rotary motion and rotational mathematics, enhancing learning where hardware access is limited. In decentralized manufacturing, it extends the capabilities of existing low-cost CNC platforms, supporting local production of parts previously restricted to centralized facilities.

### 4.5 Future Directions

Future work could explore semi-continuous rotary emulation, feedback integration, broader hardware compatibility, and AI-assisted optimization of angular steps to balance fidelity and machining time.

## 5. Conclusions

A simple Python post-processor can extend GRBL-based CNC machines with practical rotary functionality, achieving deviations within ±0.25 mm—suitable for prototyping, education, and light-duty manufacturing.

**Key contributions:**

**Accessibility and Cost Reduction:** Eliminates specialized rotary hardware or firmware modifications.

**Software-Driven Rotary Emulation:** Discrete indexing and automated G-code modification simulate rotation on standard GRBL controllers. GUI and visualization modules reduce errors and enable rapid validation.

**Educational and Decentralized Impact:** Low-cost exposure to multi-axis machining supports hands-on learning and distributed fabrication.

Limitations include minor surface artifacts, dependence on well-calibrated hardware, and lack of real-time feedback, restricting high-precision or heavy-duty applications.

Future work could reduce faceting via continuous interpolation, integrate higher-resolution motion control and adaptive feedrates, and expand hardware compatibility and web-interface usability.

This methodology provides a practical, low-cost route to extend three-axis CNC machines, demonstrating that functional multi-axis enhancements can be achieved purely via software, advancing the democratization of digital fabrication.

**Supplementary Materials:** The **Python post-processor** developed for this study is available at the following DOI: https://doi.org/10.5281/zenodo.17211808

The web-based G-code converter can be accessed at https://cnc-rotary-axis-g-code-converter-seven.vercel.app/.

No additional supplementary figures, tables, or videos are provided.

**Author Contributions:** Conceptualization, P.P.; Methodology, P.P.; Software, P.P.; Validation, P.P.; Formal analysis, P.P.; Writing—original draft preparation, P.P.; Supervision, P.P.; Project administration, P.P.; Investigation, D.L.; Writing—review and editing, D.L.; Resources, D.L.; Data curation, D.V.; Visualization, D.V. All authors have read and agreed to the published version of the manuscript.

**Funding:** This research received no external funding.

**Data Availability Statement:** All data generated or analyzed during this study are included in this published article. No additional datasets were created.

**Acknowledgments:** Author affiliations reflect the authors' academic backgrounds; the research itself was conducted independently, without institutional support or funding. The authors gratefully acknowledge Jim Fong for his independent testing of the proposed method, with results presented in Figure 10. His contributions represent insights from the CNC community through engagement on related online forums.
During manuscript preparation, the authors used ChatGPT (GPT-5) for minor grammar and style suggestions. All output was reviewed and edited by the authors, who take full responsibility for the content of the publication.

**Conflicts of Interest:** The authors declare no conflicts of interest

## Appendix A. Source Code Excerpt

The following excerpt illustrates core routines from the Python post-processor (Section 4.1), including G-code parsing, rotary synchronization, and toolpath output. Auxiliary code such as GUI layout and error handling is omitted for brevity. The full implementation is available at https://doi.org/10.5281/zenodo.17211808.

```
def calculate_passes_and_angular_displacement():
    """Calculate number of passes and angular displacement based on toolpath and
stock diameter"""
    try:
        stock_dia = float(stock_diameter_var.get())
        tool_dia = float(tool_diameter_var.get())

        # Calculate toolpath diameter (innermost radius reached by the tool)
        toolpath_dia = stock_dia - (2 * tool_dia)

        if toolpath_dia <= 0:
            messagebox.showerror("Error", "Tool diameter is too large for stock di-
ameter")
            return None, None
```

```
        # Calculate number of passes needed for complete revolution
        overlap_factor = 0.8    # Default: 80% overlap for better surface finish
        pass_width = tool_dia * overlap_factor
        num_passes = max(1, int(math.ceil(math.pi * toolpath_dia / pass_width)))

        # Calculate angular displacement per pass
        angular_displacement = 360.0 / num_passes

        # Update display variables
        num_passes_var.set(str(num_passes))
        angular_displacement_var.set(f"{angular_displacement:.2f}")

        return num_passes, angular_displacement

    except ValueError:
        return None, None
```

## Appendix B. Web Interface Access

The live web interface for the rotary G-code converter is available at: https://cnc-rotary-axis-g-code-converter-seven.vercel.app/